# A New Method of Pixel-level In-situ U-value Measurement for Building Envelopes Based on Infrared Thermography

Zihao Wang;[1] Yu Hou, Ph.D.;[2]
Lucio Soibelman, Ph.D., Dist.M.ASCE[3]

[1]Sonny Astani Department of Civil and Environmental Engineering, University of Southern California, Los Angeles, CA 90089. (corresponding author). Email: zwang293@usc.edu
[2]Assistant Professor, Department of Construction Management, Western New England University, Springfield, MA 01119. Email: yu.hou@wne.edu
[3]Professor, Sonny Astani Department of Civil and Environmental Engineering, University of Southern California, Los Angeles, CA 90089. Email: soibelma@usc.edu

## ABSTRACT

The potential energy loss of aging buildings traps building owners in a cycle of underfunding operations and overpaying maintenance costs. Energy auditors intending to generate an energy model of a target building for performance assessment may struggle to obtain accurate results as the spatial distribution of temperatures is not considered when calculating the U-value of the building envelope. This paper proposes a pixel-level method based on infrared thermography (IRT) that considers two-dimensional (2D) spatial temperature distributions of the outdoor and indoor surfaces of the target wall to generate a 2D U-value map of the wall. The result supports that the proposed method can better reflect the actual thermal insulation performance of the target wall compared to the current IRT-based methods that use a single-point room temperature as input.

## INTRODUCTION

Approximately 75% of buildings are energy inefficient, resulting in significant energy losses to maintain a comfortable indoor environment (O'Grady et al. 2017). The building envelope, which consists of external walls, windows, the roof, and the floor of a building, is the largest contributor to a building's energy loss. Approximately 50% of a typical building's total energy consumption is dissipated through the envelope as energy loss (Feng et al. 2016). According to the IEA's technology roadmap for energy-efficient building envelopes (IEA 2013), improving the energy performance of building envelopes can reduce 57% of commercial energy consumption and 42% of residential energy consumption. Therefore, researchers worldwide have conducted various studies to improve the energy performance of building envelopes.

Many existing simulation tools can provide auditors with a detailed assessment of the energy performance of building envelopes. However, there is often a significant difference between the simulated performance and the actual performance based on in-situ measurements. This is defined as the "energy performance gap" (Bayomi et al. 2021). Reducing this gap and obtaining accurate results depends on the availability of more precise inputs that reflect the building envelope's existing condition, such as thermal characteristics and ambient climate data (Mahmoodzadeh et al.

2022). Therefore, understanding the existing condition of building envelopes is critical for an accurate assessment of building energy performance.

Thermal transmittance (U-value) is one of the most important inputs for the development of energy models because it directly describes the thermal insulation performance of a building component. Heat flux meters (HFM) and IRT are the most commonly used methods to measure the U-value of building envelopes. However, it is difficult for auditors to learn the U-value distribution of building envelopes using the HFM method. HFM typically requires at least 72 hours to complete a test for one or more points (Gaspar et al. 2018). Next, IRT determines the surface temperature on a pixel-by-pixel basis by receiving and analyzing the radiation emitted from the surface. The surface's U-value can then be calculated using a calculation formula based on radiation and convection. Mounted on unmanned aircraft systems (UASs), IRT cameras can capture infrared images around the target building in a few minutes and calculate the in-situ U-value of its envelope. It can also focus on a group of buildings in a large neighborhood and efficiently detect existing thermal anomalies (Hou et al. 2021). However, IRT-based methods face challenges in obtaining accurate U-values. Current IRT-based methods typically use the room temperature measured at a single point as one of the inputs to calculate the U-value of its exterior wall. This can increase calculation errors because the spatial distribution of room air temperature is not fully accounted for. Consequently, the actual energy performance of the target envelope cannot be accurately assessed.

To address this gap, this paper introduces the preliminary results of research that proposes a new IRT-based method for the U-value measurement of building envelopes. This method uses a new formula that takes the temperature of the outdoor and indoor surfaces of the target envelope as input. The new formula considers the 2D temperature distribution of both surfaces of the envelope so that the resulting U-value reflects the actual thermal insulation performance of the envelope. Wall is used as the case in this paper.

**RELATED WORK**

Auditors typically perform quantitative IRT to quantify the energy loss and evaluate the energy performance of buildings. U-value remains the primary focus of quantitative IRT studies. Infrared cameras mounted on UASs allow auditors to calculate U-values and quantify energy loss, not only for individual components but also for an entire building or multiple buildings in a short period of time (Hou et al. 2022). UASs are well suited to take advantage of the outdoor environment, although the environment is unstable and dynamic. However, quantitative methods require a stable environment; otherwise, significant errors and uncertainties in U-value measurement can occur (Mahmoodzadeh et al. 2021, 2022). Several IRT studies have reported that the maximum deviation between the theoretical and measured U-values ranges from 12% to 73% (Tejedor et al. 2020), or even from 348% to 383% (Arjoune et al. 2021). To reduce the quantification error, many researchers have contributed their efforts. Some have optimized the U-value calculation formula (Nardi et al. 2018), while others have implemented a 2D pixel-level U-value map at the building component level to more accurately represent the U-value distribution (Tejedor et al. 2020, 2021). However, these studies are feasible only in laboratory environments to obtain accurate results.

Researchers have used a formula based on radiation and convection to calculate the U-value (Mahmoodzadeh et al. 2021, 2022; Tejedor et al. 2020, 2021). The convective heat transfer coefficient was calculated as a function of the Rayleigh number, which depends on the air-to-surface temperature difference. These studies used Eq. 6, which is recommended for calculating the natural convective heat transfer from a vertical plate (Churchill and Chu 1975). The same equation is used in this study. On the other hand, existing calculation formulas use as inputs the indoor and outdoor air temperatures measured by a single thermal sensor or calculated by averaging the measurements of several thermal sensors. The spatial distribution of air temperature near the outdoor and indoor surfaces of the target envelope (e.g., an exterior wall of a room) is not considered when calculating its U-value. Consequently, the resulting U-value may not accurately reflect the thermal insulation performance of the envelope.

In summary, ignoring the diversity of indoor air temperature distribution causes errors and uncertainties in the U-value measurement and to the subsequent performance assessment. To address this gap, this study proposes a new method for measuring the U-value of building envelopes. The proposed method uses the pixel-level outdoor and indoor surface temperatures of the target envelope to calculate the surface-to-surface temperature difference, which accurately reflects its 2D temperature distribution.

**PIXEL-LEVEL U-VALUE MEASUREMENT**

Existing methods typically use indoor and outdoor air temperatures measured by a single sensor to calculate U-values based on outdoor IRT (Mahmoodzadeh et al. 2021) and indoor IRT (Tejedor et al. 2020). However, the spatial distribution of air temperature needs to be considered, especially the indoor air temperature distribution as it is significantly affected by the interior layout and the location of heat sources. In addition, indoor air temperature distribution varies vertically as hot air rises and cold air sinks. In the IRT-based U-value measurement, the accuracy of the measured U-value depends on the air temperatures near the outdoor and indoor surfaces of the target envelope. Therefore, considering the spatial distribution of the indoor air temperature can reflect the actual surroundings of the target envelope and reduce the error in its U-value measurement.

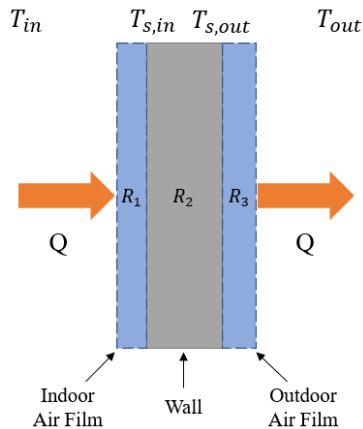

**Figure 1. Thermal Resistance Network of An Exterior Wall**

According to the heat transfer theory, the heat flux (Q) transferred through a thermal resistance network (R) is constant when the indoor environment reaches thermal equilibrium (Eq. 1). The two boundaries near both surfaces of the wall are called "air films", which needs to be considered in U-value calculation. In this study, 0.12 K·m2/W and 0.03 K·m2/W were used as the reference

R-value of indoor and outdoor air films, respectively (Rabia 2021). Then, a new U-value calculation formula for a wall can be derived based on its outdoor and indoor surface temperatures.

$$Q = \frac{T_{in} - T_{s,in}}{R_1} = \frac{T_{s,in} - T_{s,out}}{R_2} = \frac{T_{s,out} - T_{out}}{R_3} = \frac{T_{in} - T_{out}}{R_1 + R_2 + R_3} \quad (1)$$

The U-value of a wall can be calculated by dividing the sum of the convection heat transfer at the surface and the radiation exchange between the exterior wall surface and the outdoor environment by the temperature difference between indoors and outdoors. Thus, the U-value calculation formula using IRT outside the building is shown as Eq. 2. Alternatively, based on Eq. 1, the U-value of the target wall can be calculated using Eq. 3. Air films need to be considered separately.

$$U = \frac{q_r + q_c}{T_{in} - T_{out}} = \frac{\varepsilon \cdot \sigma \cdot (T_{s,out}^4 - T_{out}^4) + h_{ext} \cdot (T_{s,out} - T_{out})}{T_{in} - T_{out}} \quad (2)$$

$$U_{wall} = \frac{\varepsilon \cdot \sigma \cdot (T_{s,out}^4 - T_{out}^4) + h_{ext} \cdot (T_{s,out} - T_{out})}{T_{s,in} - T_{s,out}} \quad (3)$$

Where $\varepsilon$ is the emissivity coefficient of the element, which depends on the type of material and temperature of the surface; $\sigma$ is Stefan-Boltzmann's constant with a value of $5.67 \times 10^{-8}$; $h_{ext}$ is the exterior convective heat transfer coefficient; $T_{in}$ is the air temperature near the indoor surface of the wall; $T_{out}$ is the air temperature near the outdoor surface of the wall; $T_{s,in}$ is the indoor surface temperature of the wall; and $T_{s,out}$ denotes the outdoor surface temperature of the wall. The temperature is to be used in the unit degree Kelvin.

The exterior convective heat transfer coefficient $h_{ext}$ can be calculated using the following formula (The temperature is to be used in the unit degree Kelvin):

$$h_{ext} = \frac{Nu \cdot k}{L} \quad (4)$$

$$R_a = \frac{g \cdot \beta \cdot (T_{s,out} - T_{out}) \cdot L^3}{v^2} \cdot P_r \quad (5)$$

$$Nu = \{0.825 + \frac{0.387 \cdot R_a^{\frac{1}{6}}}{[1 + (\frac{0.492}{P_r})^{\frac{9}{16}}]^{\frac{8}{27}}}\}^2 \quad (6)$$

Where $L$ is the height of the wall seen from inside the building (m); $k$ is the thermal conductivity of air; $R_a$ is the Rayleigh number, $P_r$ is the Prandtl number, and $Nu$ is the Nusselt number. $g$ is the gravitation (9.80665 m/s2); $\beta$ is the volumetric temperature expansion coefficient, where all fluid properties should be evaluated at the film temperatures, so $\beta = \frac{1}{T_m}$ where $T_m = \frac{T_{out} + T_{s,out}}{2}$. The $v$ is the air kinematic viscosity.

# CASE STUDY

The testbed for this study is an individual building: EASI (Efficient, Affordable, Solar Innovation) House. EASI House is located at the Western New England University campus and was built in 2015. The north wall of the west bedroom was the target wall.

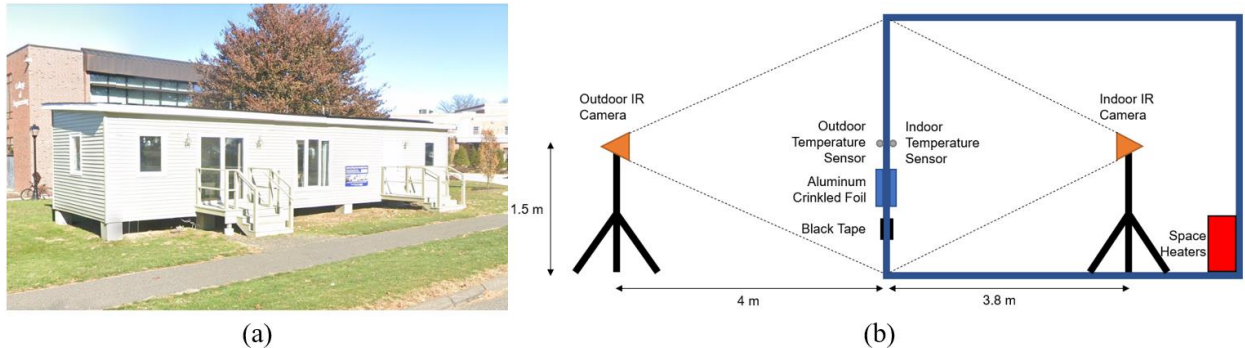

**Figure 2: (a) EASI House; (b) Experiment Setup**

The test was conducted on a cloudy day in March to avoid solar radiation. The wind speed was less than 1.0 m/s. We conducted the test with a temperature difference across the target wall of at least 15°C. The outdoor infrared camera (DJI Mavic 3T) and indoor infrared camera (FLIR DUO Pro R) were mounted on a tripod 1.5 m above the ground and positioned 4 m and 3.8 m from the target wall respectively. Infrared and RGB images were taken by the two cameras to obtain the outdoor and indoor surface temperatures of the target wall. The target wall was divided into four sections and infrared images were taken section by section because the indoor infrared camera could not capture the entire wall in the room. The wall was monitored for 15 minutes with a data acquisition interval of 2 seconds. Two space heaters were placed in the corner of the room. The heater started heating when indoor temperature was lower than 21°C and stopped heating at 22°C.

Two temperature sensors were installed in the middle of one side of both surfaces of the target wall to measure indoor and outdoor temperatures and relative humidities. A group of four temperature sensors was mounted on the indoor surface of the wall in a 2x2 matrix. The four temperature sensors were moved to the next position when the heaters resumed heating at 21°C. The indoor temperature sensor acted as a reference that guided the heaters to turn on and off. Finally, a 4x4 temperature matrix of the air near the indoor surface of the wall was generated.

A crumpled and re-flattened aluminum foil was used as the reflector to measure the reflected temperature. An electrical tape with a known emissivity of 0.95 (3 M Scotch Super 88 Vinyl) was used to measure the emissivity of the wall surface. All infrared images were calibrated by adjusting reflected temperature, wall emissivity, ambient temperature, relative humidity, and distance. Four infrared images and corresponding RGB images taken at the same reference temperature value were selected. The corresponding 2x2 temperature matrices were also selected to form a 4x4 matrix. Four groups of such images and 4x4 matrix were selected. The pixel-level U-value of the target wall was calculated using Eq. 2 and Eq. 3 with three different settings for the indoor air temperature input: **(1)** single point temperature measured by the indoor sensor; **(2)** temperatures measured by the 4x4 sensor matrix; **(3)** indoor surface temperature of the wall measured by the indoor infrared camera (the proposed method). Then, four groups of three 2D U-value maps using the three settings respectively were generated for the wall with a form of 40x30 mesh. Finally, one 2D U-value map was generated for each setting by averaging the four 2D U-value maps.

**RESULTS AND DISCUSSION**

Three 2D U-value maps based on the three settings were generated as shown in Fig. 3. All three maps clearly show the U-value distribution of the target wall. For example, the difference in U-value between insulations and studs can be clearly observed. Compared to the other two maps, the full profile of the studs inside the wall could be observed in Fig. 3(c), which reflects the thermal insulation performance of the target wall more clearly.

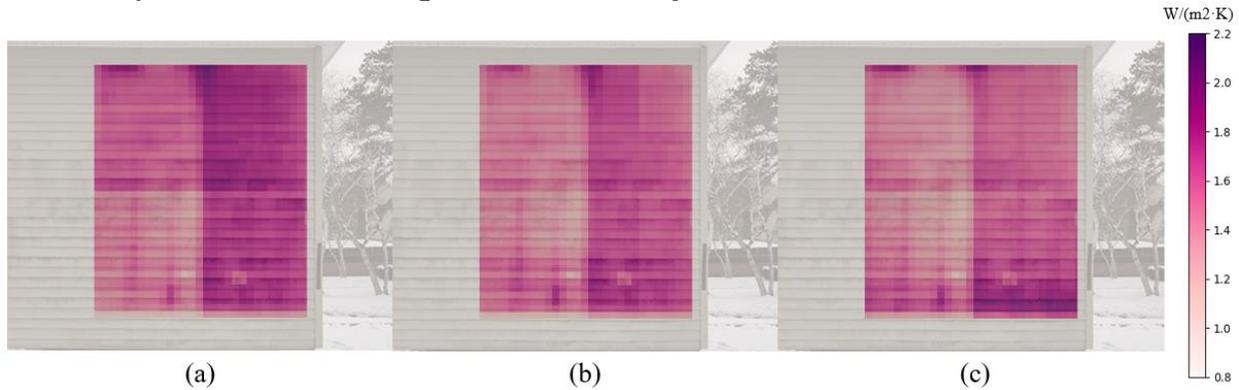

**Figure 3. U-value Map: (a) Single Point Temperature; (b) 4 x 4 Temperature Matrix; (c) Indoor Surface Temperature**

A U-value measurement HFM device (FluxTeq R-value measurement system) was used to measure the U-value of the target wall. The U-value was measured by averaging 164 measurements in the normal part of the target wall taken over a two-hour period. The average U-values of the three settings were calculated and compared with the U-value measured by the HFM device in Table 1. The result shows that the proposed method (Surface) gives a similar result to the HFM device, although the U-value measured by the HFM device may not be the actual U-value of the target wall.

**Table 1. U-value (W/(m2·K)) Measurement Comparison**

| Setting | Single Point(1) | 4x4 Matrix(2) | Surface(3) | HFM |
|---|---|---|---|---|
| Min | 0.990 | 1.040 | 1.051 | - |
| Max | 2.120 | 1.952 | 2.054 | - |
| Average | 1.626 | 1.528 | 1.504 | 1.609 |

Two maps reflecting the difference in the measured U-value between two settings were generated (Fig. 4): (a) shows the U-value difference between Setting (3) and Setting (1); (b) shows the U-value difference between Setting (3) and Setting (2). In (a), higher U-values were calculated for the bottom part of the wall using Setting (3) compared to Setting (1). This result corresponds to the fact that the air at the bottom is colder than the air in the middle. The difference in the measured U-values results from the difference in the air temperature near the indoor surface of the wall ($T_{in}$). Compared to Setting (3), Setting (1) used one temperature value as the input, resulting in lower and higher temperature differences between the outdoor and indoor surfaces of the wall (denominator of Eq. 2) at the bottom and top of the wall respectively. In (b), a similar result reinforces the above finding. In addition, the calculated U-value using Setting (3) is higher than that using Setting (2) at the edges of the wall. This observation results from the lower indoor air temperature at the edges of the room compared to the center due to airflow. Setting (2) used the same air temperature for the edge portion as the relatively central portion, causing a higher temperature difference between the outdoor and indoor surfaces of the wall.

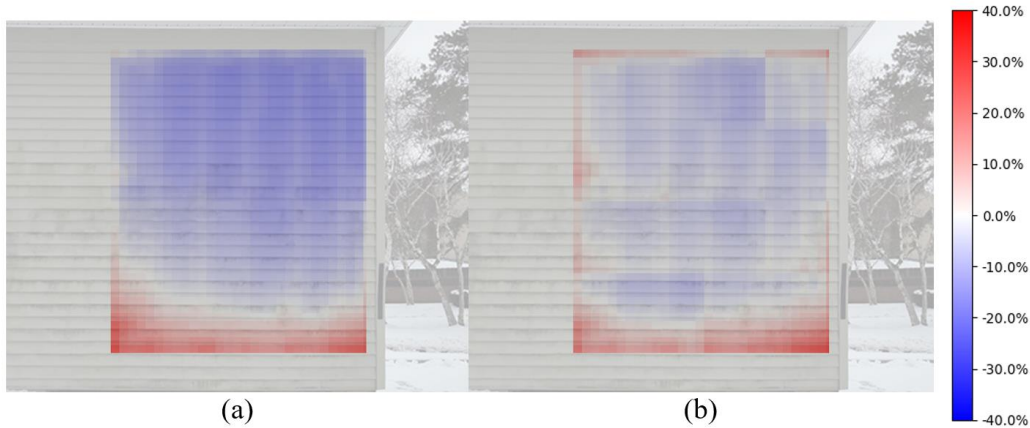

(a)           (b)

**Figure 4. U-value Difference Map: (a) Indoor Surface Temperature v.s. Single Point Temperature; (b) Indoor Surface Temperature v.s. 4 x 4 Temperature Matrix**

Some improvements can be made in future studies. The small size of the room made data collection complex. Instead, an indirect method was used to capture the thermal information of the wall. This leads to the difference in U-value measurement at the four parts of the wall in the same setting (Fig. 3). The reason for this is that IRT data collection is highly dependent on a stable environment and configurations. Future research is needed to determine appropriate configurations (e.g. collection time, camera angle) to collect information from building components to ensure a robust U-value measurement. Due to the difficulty of measuring U-values at a fine resolution (e.g. pixel-level measurement) using HFM, this study only measured the U-value of a single point using HFM. A future study is needed to validate the accuracy of the proposed method in measuring the U-value.

**CONCLUSION**

This paper proposes a new IRT-based method for measuring the U-value of building envelopes on site. The proposed method uses the outdoor and indoor surface temperatures of the target envelope to calculate the surface-to-surface temperature difference. It reflects the actual thermal insulation performance of the envelope by considering the 2D spatial temperature distributions of its outdoor and indoor surfaces. This new method helps generate an energy model for the building envelope that reflects its actual thermal insulation performance. In future work, we plan to develop a three-step framework for localizing thermal anomalies, quantifying energy losses, and generating energy models of building envelopes using the proposed method in this paper.